\documentclass[11pt]{article}

\usepackage[preprint]{acl}

\usepackage{times}
\usepackage{latexsym}

\usepackage[T1]{fontenc}

\usepackage[utf8]{inputenc}

\usepackage{microtype}

\usepackage{inconsolata}

\usepackage{graphicx}
\usepackage[table]{xcolor}
\usepackage[dvipsnames, svgnames]{xcolor}
\usepackage{pifont}
\usepackage{bbding}
\usepackage{algorithm}
\usepackage{algorithmic}
\usepackage{booktabs}
\usepackage{multirow}
\usepackage{multicol}
\usepackage{amsmath, amssymb, bm}
\usepackage{mathtools, physics}
\usepackage[most]{tcolorbox}
\usepackage{pifont}
\usepackage{makecell} 
\usepackage{enumitem}

\title{LogicPoison: Logical Attacks on Graph Retrieval-Augmented Generation}

\author{
 \textbf{Yilin Xiao\textsuperscript{$\spadesuit$}}\thanks{Equal contribution.},
 \textbf{Jin Chen\textsuperscript{$\clubsuit$}}\footnotemark[1],
 \textbf{Qinggang Zhang\textsuperscript{$\spadesuit$}},
 \textbf{Yujing Zhang\textsuperscript{$\spadesuit$}},
\\
 \textbf{Chuang Zhou\textsuperscript{$\spadesuit$}},
 \textbf{Longhao Yang\textsuperscript{$\clubsuit$}},
 \textbf{Lingfei Ren\textsuperscript{$\clubsuit$}}\thanks{Corresponding author.},
 \textbf{Xin Yang \textsuperscript{$\clubsuit$}},
 \textbf{Xiao Huang\textsuperscript{$\spadesuit$}}
\\
 \textsuperscript{$\clubsuit$}Southwestern University of Finance and Economics, 
 \textsuperscript{$\spadesuit$}The Hong Kong Polytechnic University \\
 \texttt{yilin.xiao@connect.polyu.hk} \hspace{1.8em}
 \texttt{jordchen@163.com} \\
 \texttt{renlf@swufe.edu.cn} \hspace{0.9em} \texttt{xiaohuang@comp.polyu.edu.hk} \\
}

\begin{document}
\maketitle
\begin{abstract}
Graph-based Retrieval-Augmented Generation (GraphRAG) enhances the reasoning capabilities of Large Language Models (LLMs) by grounding their responses in structured knowledge graphs. Leveraging community detection and relation filtering techniques, GraphRAG systems demonstrate inherent resistance to traditional RAG attacks, such as text poisoning and prompt injection. However, in this paper, we find that the security of GraphRAG systems fundamentally relies on the topological integrity of the underlying graph, which can be undermined by implicitly corrupting the logical connections, without altering surface-level text semantics. To exploit this vulnerability, we propose \textsc{LogicPoison}, a novel attack framework that targets logical reasoning rather than injecting false contents. Specifically, \textsc{LogicPoison} employs a type-preserving entity swapping mechanism to perturb both global logic hubs for disrupting overall graph connectivity and query-specific reasoning bridges for severing essential multi-hop inference paths. This approach effectively reroutes valid reasoning into dead ends while maintaining surface-level textual plausibility. Comprehensive experiments across multiple benchmarks demonstrate that \textsc{LogicPoison} successfully bypasses GraphRAG's defenses, significantly degrading performance and outperforming state-of-the-art baselines in both effectiveness and stealth. Our code is available at \textcolor{blue}{\url{https://github.com/Jord8061/logicPoison}}.
\end{abstract}

\section{Introduction}
Retrieval-Augmented Generation (RAG)~\cite{RAG,gao2023retrieval,LAG} has emerged as a promising paradigm for enhancing Large Language Models (LLMs)~\cite{gpt,Claude,deepseek} by leveraging external knowledge bases. However, existing RAG systems face significant challenges when processing large-scale, unstructured corpora in real-world scenarios. The relevant information is often unevenly distributed across heterogeneous documents~\cite{mybench,yang2026graph}. Consequently, the contexts retrieved by traditional RAG are frequently voluminous, fragmented, and lacking in structural organization, leading to inconsistencies in generation accuracy and coherence. To address these limitations, Graph Retrieval-Augmented Generation (GraphRAG)~\cite{zhang2025survey,peng2024graph,RRP} has gained prominence as a robust solution. By structuring knowledge into graphs, GraphRAG models hierarchical relationships and supports efficient multi-hop retrieval and reasoning, enabling more reliable use of background knowledge.

\begin{figure}[t]
    \centering
    \includegraphics[width=1\linewidth]{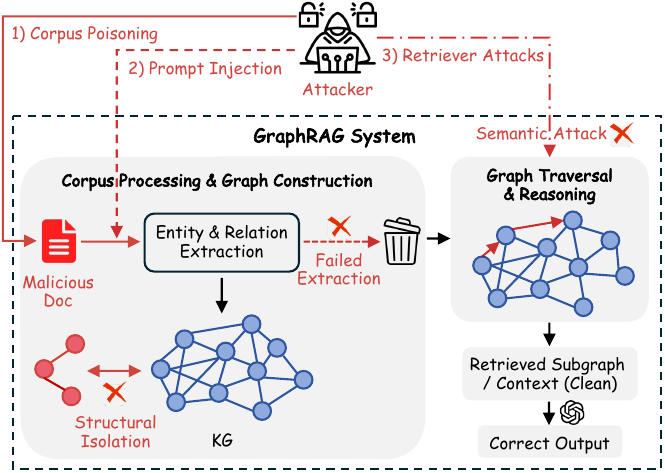}
    \caption{Traditional attacks on LLM or RAG are difficult to pose an effective threat to GraphRAG system. This is mainly due to two characteristics. First, the construction process of the knowledge graph naturally filters part of the attack content. Second, the unique retrieval and reasoning mechanism of graph structure further improves the difficulty of attack implementation.}
    \label{fig:intro}
\end{figure}


Security remains a critical challenge for Retrieval-Augmented Generation (RAG) systems due to their reliance on external corpora, which exposes them to adversarial attacks. As shown in Figure~\ref{fig:intro}, prior research has identified three primary attack types: (i) Corpus Poisoning~\cite{poisonedrag}: injecting malicious documents into the knowledge source; (ii) Prompt Injection~\cite{Defending, Signed-prompt}, where adversarial instructions are embedded within retrieved text to hijack model behavior; and (iii) Retriever Attacks~\cite{universalTriggers,Shafran}, that craft deceptive queries to degrade the retrieval accuracy.

However, these methods are largely ineffective against GraphRAG due to its graph-based architecture. By constructing a knowledge graph from source corpora, GraphRAG introduces inherent defensive mechanisms. First, poisoning attacks are topologically marginalized, as synthetic entities fail to integrate into coherent graph communities. Second, and most critically, prompt injection is structurally prevented: retrieval operates over entity-relation subgraphs rather than raw text, thereby decoupling knowledge access from instruction execution and filtering out embedded adversarial commands. Besides that, because reasoning occurs over multiple connected paths within the graph, attacks targeting single nodes or local content fail to corrupt the system's global inference capability, as alternative uncontaminated reasoning paths remain available. Recent work like GRAGPOISON~\cite{Fire} has taken initial steps to attack GraphRAG. However, it relies heavily on LLM-generated contents to fabricate and amplify spurious relations, which is highly detectable.
To make it clear, we conduct an in-depth analysis (in Appendix ~\ref{appendix:Analysis}) on the graph-based architecture and reveal that GraphRAG functions less as a static knowledge store and more as a dynamic logical reasoning engine. Its efficacy stems not from the quantity of stored triples, but from its ability to conduct logical propagation via topological structures. Consequently, defeating GraphRAG requires disrupting its logical reasoning chains rather than merely corrupting isolated data points. However, the underlying graph structure is typically opaque (black-box) in practice, and different GraphRAG implementations employ diverse graph construction and reasoning mechanisms. This imposes two primary challenges: (i) \textit{How to achieve effective, and low-magnitude poisoning without knowledge of the underlying graph structure?} (ii) \textit{How can such an attack remain robust and transferable across diverse GraphRAG architectures?}

To overcome these challenges, we propose a novel attack framework: \textsc{LogicPoison}. Different from traditional attack methods that rely on additive noise or conflicting evidence, \textsc{LogicPoison} fundamentally targets the topological integrity of the knowledge graph constructed by GraphRAG systems. Specifically, it employs a Type-Preserving Entity Swapping mechanism to implicitly corrupt the graph structure without disrupting the textual surface. We devise a dual-pronged strategy to select target entities: (i) Global Logic Poison, which identifies and perturbs high-frequency hub nodes to shatter global graph connectivity; and (ii) Query-Centric Logic Poison, which utilizes Chain-of-Thought extraction to pinpoint and sever specific reasoning bridges essential for multi-hop queries. By applying cyclic permutations to these key entities within their respective types, we ensure that while the document remains semantically readable, the underlying graph topology is rerouted into logical dead ends or spurious shortcuts. Our contributions are summarized as follows:

\begin{itemize}
    \item We identify a critical vulnerability in GraphRAG systems: while they are robust to unstructured noise, they are highly fragile to structural logic corruption. We formalize this as a topological security problem, shifting the attack surface from poisoned content injection to logic rewiring.
    \item We propose \textsc{LogicPoison}, a unified framework that combines global centrality analysis with query-specific reasoning extraction. This method allows for low-magnitude, stealthy modifications that effectively bypass the community summarization and filtering mechanism defenses inherent in GraphRAG.
    \item We conduct comprehensive experiments on multiple benchmarks across various SOTA GraphRAG architectures and base LLMs. The results demonstrate that \textsc{LogicPoison} significantly degrades GraphRAG performance, outperforming existing baselines.
\end{itemize}

\begin{figure*}[t]
    \centering
    \includegraphics[width=1.0\linewidth]{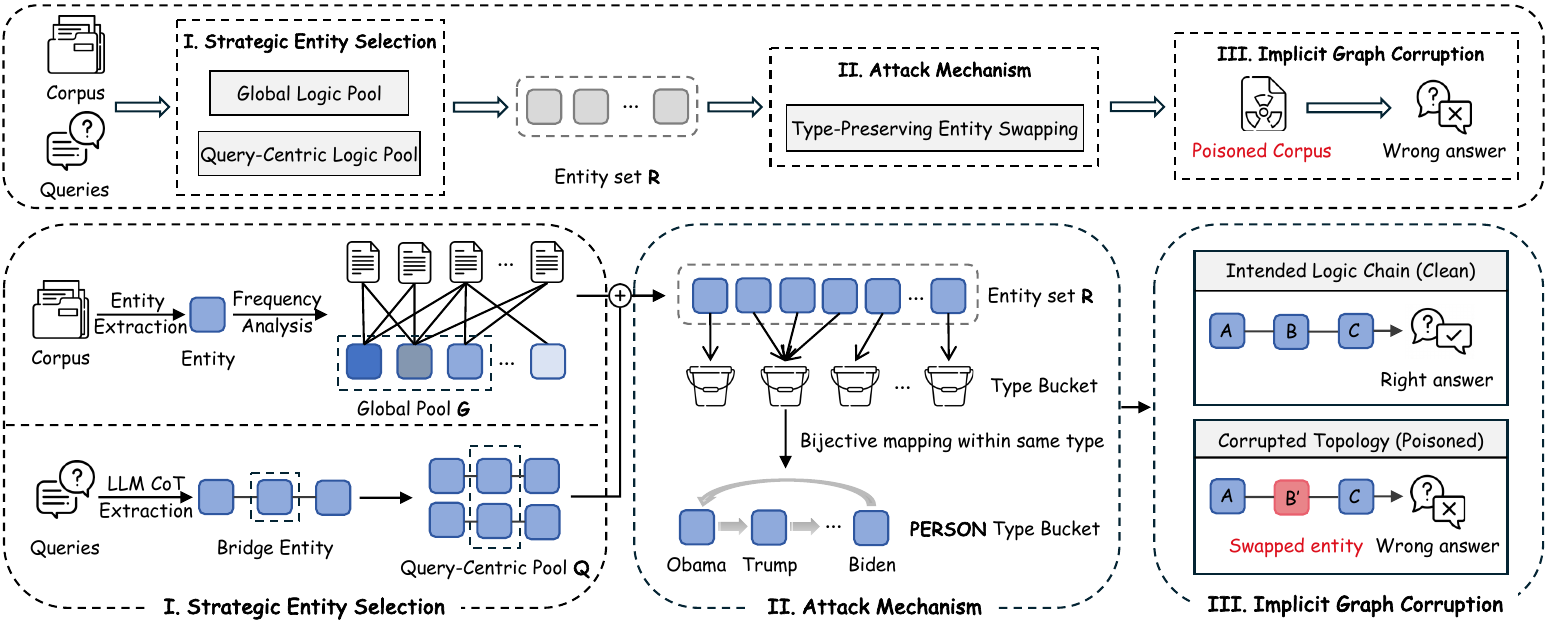}
    \caption{The overall framework of \textsc{LogicPoison}. The attack pipeline is divided into three stages: \textbf{I. Strategic Entity Selection}, where target entities are identified via a dual-pronged strategy combining global logic hubs and query-centric reasoning bridges into a unified set $\mathcal{R}$. \textbf{II. Attack Mechanism}, which employs a type-preserving cyclic permutation to swap entities within their respective type buckets in the corpus. \textbf{III. Implicit Graph Corruption}, demonstrating how the poisoned corpus subtly rewires the implicit topology of the constructed graph, severing valid reasoning chains ($A \to B \to C$) and re-routing them to incorrect entities ($A \to B' \to C$), leading to wrong answers while maintaining textual plausibility.}
    \label{fig:framework}
\end{figure*}

\section{Related Work}


The security of Retrieval-Augmented Generation models has gained considerable attention in the research community in recent years. The concept of Universal Adversarial Triggers~\cite{universalTriggers} and Indirect Prompt Injection~\cite{Defending, Signed-prompt} served as the foundation for initial results that pointed out the vulnerability of large language models to any type of input attack. Starting with this point, various attacks on the RAG models have emerged in the pipeline: PoisonedRAG~\cite{poisonedrag}defined knowledge-base poisoning for directing model behavior; PANDORA~\cite{Pandora} utilized retrieval functionality for conducting indirect jailbreak attacks; TrojanRAG~\cite{TrojanRAG} optimized joint attacks for inserting backdoors in the retrieval contexts; Jamming attacks~\cite{Shafran} initiated denial-of-service conditions by promoting blocking documents; besides, privacy leakage issues intrinsic to the retrieval process were uncovered by further studies~\cite{zeng-good}. Most recently, GRAGPOISON~\cite{Fire} generalized GraphRAG attacks by injecting texts written by LLMs for simulating fake linkages in the graphs. The existing techniques were based on explicit content injection and additive perturbation. This work introduces \textsc{LogicPoison}, which changes the paradigm by re-modeling the topology of logical graphs to effectively manipulate GraphRAG by stealthy, type-preserving swapping, which corrupts GraphRAG in a logically implicit manner over more obvious generations.

\section{Preliminaries}
Given a retrieval corpus $\mathcal{C} = \{d_i\}_{i=1}^N$ and a set of queries $\mathcal{Q}$. A GraphRAG system builds an implicit or explicit knowledge graph $\mathcal{G} = (\mathcal{V}, \mathcal{E})$ from $\mathcal{C}$ to perform reasoning. Our objective is to obtain a poisoned corpus $\tilde{\mathcal{C}}$ via a perturbation function $\mathcal{T}: \mathcal{C} \to \tilde{\mathcal{C}}$, such that the system's reasoning on $\tilde{\mathcal{C}}$ is disrupted, while being imperceptible and grammatically coherent at the text level. In networked systems, the effectiveness of poisoning attacks is closely related to the structural importance of the targeted nodes. Since knowledge graphs are not available before graph construction, classical topologically centrality measures cannot be applied directly. In such cases, corpus-level statistics offer alternative insights for node importance. From a spectral graph theory perspective~\cite{spielman2012spectral}, global connectivity and information propagation are largely determined by the dominant modes of the adjacency structure, which are disproportionately influenced by high-degree nodes (Check more details in Appendix \ref{sec:spectral}). Consequently, high-frequency entities implicitly control the backbone of information flow, making them particularly effective targets for poisoning attacks.

\section{The Framework of LogicPoison}
\label{sec:method}
We propose the \textsc{LogicPoison}, a framework specifically targeting the GraphRAG systems on the principle of topological disruption. Contrary to traditional poisoning methods that inject conspicuous amounts of misinformation, our solution carefully rewrites the \emph{logical reasoning chains} inherent in the corpus. By strategically permuting entities, we induce semantically plausible yet logically fallacious paths in the constructed knowledge graph, misleading the system into generating confident but incorrect answers.

The corrupted corpus $\tilde{\mathcal{C}}$ can be built by performing the operation of type-preserving entity swapping on a candidate set of entities $\mathcal{S}$. In order to have structural impact and querying relevance at the same time,  we compose $\mathcal{S}$ using two complementary sub-strategies:

\begin{itemize}
    \item \textbf{Global Logic Poison:} Targets corpus-level \emph{logic hubs} to distort the global connectivity of the knowledge graph.
    \item \textbf{Query-Centric Logic Poison:} Targets specific entities essential for multi-hop logical reasoning chains to ensure the attack covers the precise paths required by user queries.
\end{itemize}

\subsection{Global Logic Poison}
GraphRAG systems rely on central entities to connect disparate information chunks. Disrupting these hubs destabilizes the global graph structure.

\paragraph{Entity Extraction.}
We treat each document $d_i \in \mathcal{C}$ as a processing knowledge unit. To identify potential knowledge graph nodes, we employ a lightweight Named Entity Recognition (NER) model (e.g., spaCy). Formally, for each document $d_i$, we extract a set of typed entities:
\begin{equation}
    E(d_i) = \{ (e, \tau) \mid e \in \mathcal{V}_{\text{vocab}}, \tau \in \mathcal{T} \},
\end{equation}
where $e$ denotes the surface form and $\tau$ represents the entity type (e.g., \textsc{Person}, \textsc{Org}, \textsc{Date}).

\paragraph{Corpus-Level Frequency Analysis.}
We approximate the topological importance of an entity by its document frequency. For an entity $e$ of type $\tau$, we define its frequency as:
\begin{equation}
    f(e, \tau) = \sum_{i=1}^{N} \mathbb{I}\big[ (e,\tau) \in E(d_i) \big],
\end{equation}
where $\mathbb{I}[\cdot]$ is the indicator function. Intuitively, entities with high $f(e, \tau)$ act as implicit hubs that bridge numerous communities in the graph.

\paragraph{Global Replacement Pool.}
To maximize disruption, we rank entities within each type $\tau$ by $f(e, \tau)$ and select the top-$n\%$ as targets:
\begin{equation}\small
    \mathcal{G}_\tau = \left\{ e \;\middle|\; \text{type}(e)=\tau \land \text{rank}(f(e,\tau)) \le n\% \right\}.
\end{equation}
The global candidate set is defined as $\mathcal{G} = \bigcup_{\tau \in \mathcal{T}} \mathcal{G}_\tau$. Poisoning these entities distorts the backbone of the reasoning graph, affecting a broad spectrum of retrieval operations.

\subsection{Query-Centric Logic Poison}
Global hubs may not cover long-tail entities critical for specific queries. To enforce path coverage, we introduce a query-guided selection mechanism.

\paragraph{Chain-of-Thought Entity Extraction.}
For each multi-hop query $q \in \mathcal{Q}$, we prompt an LLM to perform Chain-of-Thought (CoT) reasoning to identify entities essential for deriving the answer (e.g., bridge entities, comparison targets). This yields a structured set of reasoning entities:
\begin{equation}
    E_q = \{ (e, \tau, r) \},
\end{equation}
where $r$ denotes the reasoning role (e.g., \texttt{bridge}, \texttt{target}). This ensures we target logical dependencies of the query rather than superficial mentions.

\paragraph{Corpus Verification.}
To prevent the hallucination of non-existent targets, we perform a filtering step on $E_q$ against the corpus inventory:
\begin{equation}\small
    E_q^{\mathrm{corpus}} = \{ (e,\tau) \mid (e,\tau,r) \in E_q \land f(e,\tau) > 0 \}.
\end{equation}
This ensures that the attack only modifies entities that actually exist in $\mathcal{C}$ and participate in the knowledge graph construction pipeline.

\paragraph{Query-Centric Replacement Pool.}
We aggregate these verified entities across all queries to form the query-centric replacement pool:
\begin{equation}\small
    \mathcal{Q}_\tau = \bigcup_{q \in \mathcal{Q}} \{ e \mid (e,\tau) \in E_q^{\mathrm{corpus}} \}, \quad \mathcal{Q} = \bigcup_{\tau \in \mathcal{T}} \mathcal{Q}_\tau.
\end{equation}
This design guarantees that if a GraphRAG system correctly follows the intended reasoning chain, it will inevitably traverse a poisoned node.

\subsection{Type-Preserving Entity Swapping}
\label{sec:swapping}

\paragraph{Unified Candidate Set.}
We combine the global and query-centric pools into a unified set of target entities, which play complementary roles:
\begin{equation}
    \mathcal{R}_\tau = \mathcal{G}_\tau \cup \mathcal{Q}_\tau, \quad \mathcal{R} = \bigcup_{\tau \in \mathcal{T}} \mathcal{R}_\tau.
\end{equation}
Crucially, we maintain strict type constraints during swapping to ensure the poisoned text remains locally plausible and grammatically valid, thereby bypassing syntax-based filters.

\begin{table*}[t]
\centering
\resizebox{\textwidth}{!}{ 
\begin{tabular}{lllcccccccc}
\toprule
\multirow{2}{*}{\textbf{Dataset}} & \multirow{2}{*}{\textbf{LLM}} & \multirow{2}{*}{\textbf{Attack}} & \multicolumn{2}{c}{\textbf{Naive RAG}} & \multicolumn{2}{c}{\textbf{GraphRAG}} & \multicolumn{2}{c}{\textbf{GFM-RAG}} & \multicolumn{2}{c}{\textbf{HippoRAG2}} \\ 
\cmidrule(lr){4-5} \cmidrule(lr){6-7} \cmidrule(lr){8-9} \cmidrule(lr){10-11}
& & & ASR & ASR-G & ASR & ASR-G & ASR & ASR-G & ASR & ASR-G \\ 
\midrule

\multirow{6}{*}{HotpotQA} & \multirow{2}{*}{GPT-4o-mini} & PoisonedRAG & 61.8 & 72.8 & 66.8 & 80.0 & 71.6 & 70.4 & 68.0 & 66.0 \\
& & LogicPoison & \textbf{92.0} & \textbf{91.6} & \textbf{78.4} & \textbf{92.2} & \textbf{81.0} & \textbf{78.0} & \textbf{73.6} & \textbf{66.2} \\
\cmidrule{2-11}
& \multirow{2}{*}{Llama-3.1-8B} & PoisonedRAG & 51.0 & 58.8 & 64.4 & 78.0 & 55.8 & 49.4 & 68.4 & 63.6 \\
& & LogicPoison & \textbf{88.2} & \textbf{86.6} & \textbf{79.2} & \textbf{91.2} & \textbf{80.6} & \textbf{83.8} & \textbf{72.6} & \textbf{73.0} \\
\cmidrule{2-11}
& \multirow{2}{*}{Qwen-3-32B} & PoisonedRAG & 56.8 & 68.2 & 65.8 & 80.2 & 71.6 & 35.0 & 71.8 & 67.6 \\
& & LogicPoison & \textbf{85.0} & \textbf{83.8} & \textbf{84.2} & \textbf{92.0} & \textbf{76.6} & \textbf{76.8} & \textbf{76.6} & \textbf{69.8} \\
\midrule

\multirow{6}{*}{2Wiki} & \multirow{2}{*}{GPT-4o-mini} & PoisonedRAG & 64.4 & 83.4 & 58.6 & 77.4 & 57.2 & 56.6 & 74.8 & 70.4 \\
& & LogicPoison & \textbf{90.0} & \textbf{91.6} & \textbf{78.4} & \textbf{95.6} & \textbf{71.6} & \textbf{76.0} & \textbf{82.8} & \textbf{71.8} \\
\cmidrule{2-11}
& \multirow{2}{*}{Llama-3.1-8B} & PoisonedRAG & 61.4 & 77.8 & 54.8 & 73.2 & 42.2 & 37.2 & 74.8 & 70.0 \\
& & LogicPoison & \textbf{95.4} & \textbf{94.4} & \textbf{78.8} & \textbf{95.0} & \textbf{74.2} & \textbf{85.6} & \textbf{77.4} & \textbf{74.2} \\
\cmidrule{2-11}
& \multirow{2}{*}{Qwen-3-32B} & PoisonedRAG & 63.0 & 82.4 & 58.0 & 77.0 & 58.8 & 37.6 & 75.8 & 73.6 \\
& & LogicPoison & \textbf{82.2} & \textbf{83.4} & \textbf{78.6} & \textbf{93.2} & \textbf{71.4} & \textbf{79.8} & \textbf{87.4} & \textbf{76.6} \\
\midrule

\multirow{6}{*}{MuSi} & \multirow{2}{*}{GPT-4o-mini} & PoisonedRAG & 64.0 & 77.8 & 59.6 & 77.6 & 70.0 & 74.8 & 67.2 & 68.0 \\
& & LogicPoison & \textbf{99.2} & \textbf{99.2} & \textbf{91.4} & \textbf{97.0} & \textbf{92.6} & \textbf{93.4} & \textbf{90.2} & \textbf{88.0} \\
\cmidrule{2-11}
& \multirow{2}{*}{Llama-3.1-8B} & PoisonedRAG & 58.2 & 64.4 & 60.2 & 75.4 & 43.0 & 42.6 & 59.4 & 57.8 \\
& & LogicPoison & \textbf{97.4} & \textbf{97.4} & \textbf{91.6} & \textbf{97.6} & \textbf{95.0} & \textbf{98.2} & \textbf{88.8} & \textbf{92.2} \\
\cmidrule{2-11}
& \multirow{2}{*}{Qwen-3-32B} & PoisonedRAG & 66.2 & 80.0 & 61.2 & 77.6 & 59.2 & 62.0 & 67.4 & 67.0 \\
& & LogicPoison & \textbf{94.6} & \textbf{95.8} & \textbf{93.0} & \textbf{95.8} & \textbf{87.6} & \textbf{92.6} & \textbf{91.2} & \textbf{91.6} \\
\bottomrule
\end{tabular}
}
\caption{Attack Performance (ASR and ASR-GPT) of \textsc{LogicPoison} and PoisonedRAG across different RAG frameworks, Datasets, and LLMs.}
\label{tab:attack_results}
\end{table*}

\paragraph{Cyclic Permutation.}
For each entity type $\tau$, we organize the entities in $\mathcal{R}_\tau$ into a deterministic sequence $\mathbf{S}_\tau = [ e_1, e_2, \dots, e_{m} ]$. We define a cyclic permutation function $\pi_\tau$:
\begin{equation}
    \pi_\tau(e_i) = 
    \begin{cases}
        e_{i-1} & \text{if } 1 < i \le m, \\
        e_{m} & \text{if } i = 1.
    \end{cases}
\end{equation}
This mapping ensures that (i) every target entity is replaced by another entity of the same type, and (ii) the mapping is bijective and invertible.

\paragraph{Corpus Rewriting and Implicit Graph Corruption.}
Finally, we apply the perturbation function to the entire corpus. At the surface form level, any mention of $e \in \mathcal{R}_\tau$ for each document $d_i$ is substituted with $\pi_\tau(e)$. We explicitly do not modify the ground-truth answers or queries. Consequently, while the corpus $\tilde{\mathcal{C}}$ remains readable, the logical links supporting the ground truth are severed. When a GraphRAG system builds its index on $\tilde{\mathcal{C}}$ and performs chunking, embedding, and graph extraction, it implicitly constructs a corrupted topology where valid reasoning paths are rerouted to incorrect entities, misleading the reasoning process.

\section{Experiments}
\subsection{Experimental Setting}
\paragraph{Dataset.}To evaluate the efficacy of \textsc{LogicPoison} in real scenarios, we align our experimental setup with established protocols in SOTA GraphRAG research \cite{hipporag, hippo2}. We employ three widely adopted multi-hop QA benchmarks: HotpotQA \cite{hotpotqa}, 2WikiMultihopQA \cite{2wiki}, and MuSiQue \cite{musi}. These datasets represent a wide range of reasoning complexity, ranging from moderate to highly challenging tasks, and require various graph-based reasoning abilities such as bridging entities and comparative analysis. Following standard practices, we randomly sample 500 queries from the validation set of each benchmark for evaluation, while strictly align the corpus and ground-truth defined in previous work.

\paragraph{Baseline.} We benchmark \textsc{LogicPoison} against PoisonedRAG, the state-of-the-art attack for RAG systems, as well as widely adopted adversarial methods for LLMs, including Naive Attack, Prompt Injection~\cite{injection}, GCG Attack~\cite{GCG}, and Disinformation~\cite{Disinformation}. The details of all baselines are provided in Appendix ~\ref{appendix_baselines}.
\paragraph{GraphRAG methods.} To verify the robustness and universality of our attack across diverse architectures, we conduct evaluations on three representative GraphRAG frameworks: (1) Microsoft GraphRAG \cite{graphrag}, the pioneering implementation that popularized graph-based global retrieval; and (2) HippoRAG 2\cite{hippo2} and GFM-RAG \cite{gfm}, which represent the current SOTA in the field. This selection allows us to assess vulnerabilities across different graph construction and reasoning paradigms. Details of each method are provided in the Appendix.
\paragraph{Metrics.}Consistent with prior work \cite{poisonedrag}, we report the Attack Success Rate (ASR) based on substring inclusion. An attack is deemed successful if the ground truth string is absent from the model's output. Recognizing that rigid string matching fails to capture semantic nuances (e.g., synonyms or structural changes), we additionally propose ASR-GPT. This metric leverages an LLM-as-a-judge paradigm to assess semantic equivalence. Given the triplet of (question, ground truth, prediction), the judge evaluates if the prediction conveys the same meaning as the ground truth.
\paragraph{Implementation Details.} All experiments were executed on the GeForce RTX 5090. For all methods, we use Facebook/contriever as the embedding model. For the retrieval top-k parameters across different RAG methods, we set k = 10. For the top-$n\%$ of targeted entities, we set $n\%=5\%$. We use three different LLMs such as GPT-4o-mini~\cite{gpt}, Llama-3.1-8B~\cite{llama}, and Qwen-3-32B~\cite{qwen} as base LLMs to verify the generality of the experimental results.

\begin{table}[h] 
\centering
\small 
\setlength{\tabcolsep}{4pt} 
\resizebox{\linewidth}{!}{ 
\begin{tabular}{llcccc}
\toprule
\multirow{2}{*}{\textbf{LLM}} & \multirow{2}{*}{\textbf{Attack Method}} & \multicolumn{2}{c}{\textbf{GraphRAG}} & \multicolumn{2}{c}{\textbf{GFM-RAG}} \\ 
\cmidrule(lr){3-4} \cmidrule(lr){5-6}
& & ASR & ASR-G & ASR & ASR-G \\ 
\midrule

\multirow{5}{*}{GPT-4o-mini} 
& Naive Attack     & 18.8 & 15.6 & 18.8 & 10.0 \\
& Prompt Injection & 58.8 & 80.4 & 38.2 & 47.6 \\
& GCG Attack       & 19.6 & 21.6 & 19.2 & 9.6 \\
& Disinformation   & 54.6 & 75.0 & 42.0 & 51.4 \\
\rowcolor{gray!10}& LogicPoison & \textbf{78.4} & \textbf{95.6} & \textbf{71.6} & \textbf{76.0} \\
\midrule

\multirow{5}{*}{Llama-3.1-8B} 
& Naive Attack     & 18.0 & 21.4 & 20.8 & 13.6 \\
& Prompt Injection & 57.0 & 78.2 & 30.4 & 32.6 \\
& GCG Attack       & 19.8 & 16.8 & 20.4 & 12.4 \\
& Disinformation   & 53.6 & 70.4 & 42.4 & 43.2 \\
\rowcolor{gray!10}& LogicPoison & \textbf{78.8} & \textbf{95.0} & \textbf{74.2} & \textbf{85.6} \\
\midrule

\multirow{5}{*}{Qwen-3-32B} 
& Naive Attack     & 19.0 & 17.2 & 19.4 & 7.2 \\
& Prompt Injection & 59.0 & 83.4 & 50.0 & 64.0 \\
& GCG Attack       & 19.6 & 20.0 & 19.2 & 7.6 \\
& Disinformation   & 54.4 & 74.0 & 50.4 & 52.8 \\
\rowcolor{gray!10}& LogicPoison & \textbf{78.6} & \textbf{93.2} & \textbf{71.4} & \textbf{79.8} \\
\bottomrule
\end{tabular}}
\caption{Attack Performance of \textsc{LogicPoison} and LLM attack methods on the 2Wiki Dataset across different LLMs and RAG frameworks.}
\label{tab:llm_attack_results}
\end{table}

\subsection{Main Results}
Table \ref{tab:attack_results} summarizes the attack performance. \textsc{LogicPoison} achieves SOTA ASR and ASR-G across all evaluated scenarios, significantly outperforming PoisonedRAG. The advantage is most evident in graph-centric systems like GraphRAG and GFM-RAG, where our topological rewiring strategy effectively dismantles the logical connectivity essential for retrieval. While HippoRAG2 shows relative resilience due to its semantic safeguards, our method still achieves a higher success rate. Moreover, in scenarios requiring deep reasoning (MuSiQue), \textsc{LogicPoison} demonstrates dominant performance, verifying that logical corruption is more potent than content injection for multi-hop tasks. Surprisingly, this advantage persists even in Naive RAG, indicating the broad applicability of entity-level logic poisoning.

We further compare the effect difference between \textsc {LogicPoison} and the traditional LLM attack method. Table ~\ref{tab:llm_attack_results} shows that \textsc {LogicPoison} shows significant advantages in different LLM models and GraphRAG methods. Specifically, Naive Attack and GCG Attack have poor performance. The core reason is that Naive texts and text suffixes are easy to be filtered in the graph construction stage. Prompt Injection has a certain attack effect. The triple extraction process from GraphRAG usually relies on LLM, which may perform wrong triple extraction operations under the influence of injection prompt words. Disinformation has the possibility of being included in the Knowledge Graph by generating false information highly relevant to the query. However, on the whole, these traditional methods are far less effective than \textsc {LogicPoison}, because they focus on unstructured text injection which is not designed for the topology destruction of KG. In addition, we also find that GFM-RAG shows a stronger defense ability against various attacks than Microsoft GraphRAG, which is because GFM regards KG as an intermediate state requiring inference and denoising, and its performance depends more on the pre-trained GNN reasoning module.

\subsection{Efficiency Analysis}

\begin{table}[h]
    \centering
    \small 
    \setlength{\tabcolsep}{9.pt} 
    \begin{tabular}{lccc}
    
        \toprule
        \multirow{2}{*}{\textbf{Method}} & \textbf{Time} & \textbf{Token} & \textbf{Injected} \\
         & \textbf{Cost(s)}  & \textbf{Cost}  & \textbf{Tokens} \\
        \midrule
        PoisonedRAG & 6607.3 & 593.6 & 296813 \\
        \midrule
        w/ Global & 125.0 & 0 & 0 \\
        w/ Query-Centric & 1281.4 & 74.9 & 0 \\ \midrule
        \rowcolor{gray!10}\textbf{\textsc{LogicPoison}} & \textbf{1406.4} & \textbf{74.9} & \textbf{0} \\
        \bottomrule
    \end{tabular}
        \caption{\textbf{Efficiency Analysis.} Comparison of computational time, token cost per query, and the length of injected content in the corpus between baselines and our variants. All results are averaged over the three datasets.}
    \vspace{-5mm}
    
    \label{tab:efficiency_analysis}
\end{table}

In order to compare the efficiency of \textsc{LogicPoison} and PoisonedRAG, we evaluate it from three dimensions of time consumption, Token consumption and injection Token on three datasets. The results show that \textsc{LogicPoison} significantly outperforms the baseline method in all dimensions. Specifically, in terms of time consumption, Poisonedrag relies heavily on the large-scale generation of LLM, which leads to its slow speed. The global phase of \textsc {LogicPoison} uses lightweight NER technology, which takes very low time. Although the query-centric part needs to extract bridge entities through LLM and consumes a certain amount of time, the overall time is still 4 times faster than Poisonedrag. In terms of Token consumption, \textsc {LogicPoison} consumes only 1/8 of Poisonedrag, which significantly reduces the resource overhead. In the dimension of injection Tokens, Poisonedrag inserts massive attack-related content into a single dataset, which makes it easy to be detected. \textsc{LogicPoison} eliminates the need to inject spurious content into the dataset, thus achieving high concealment.

\subsection{Ablation Study}
\begin{figure*}[t]
    \centering
    \includegraphics[width=0.95\linewidth]{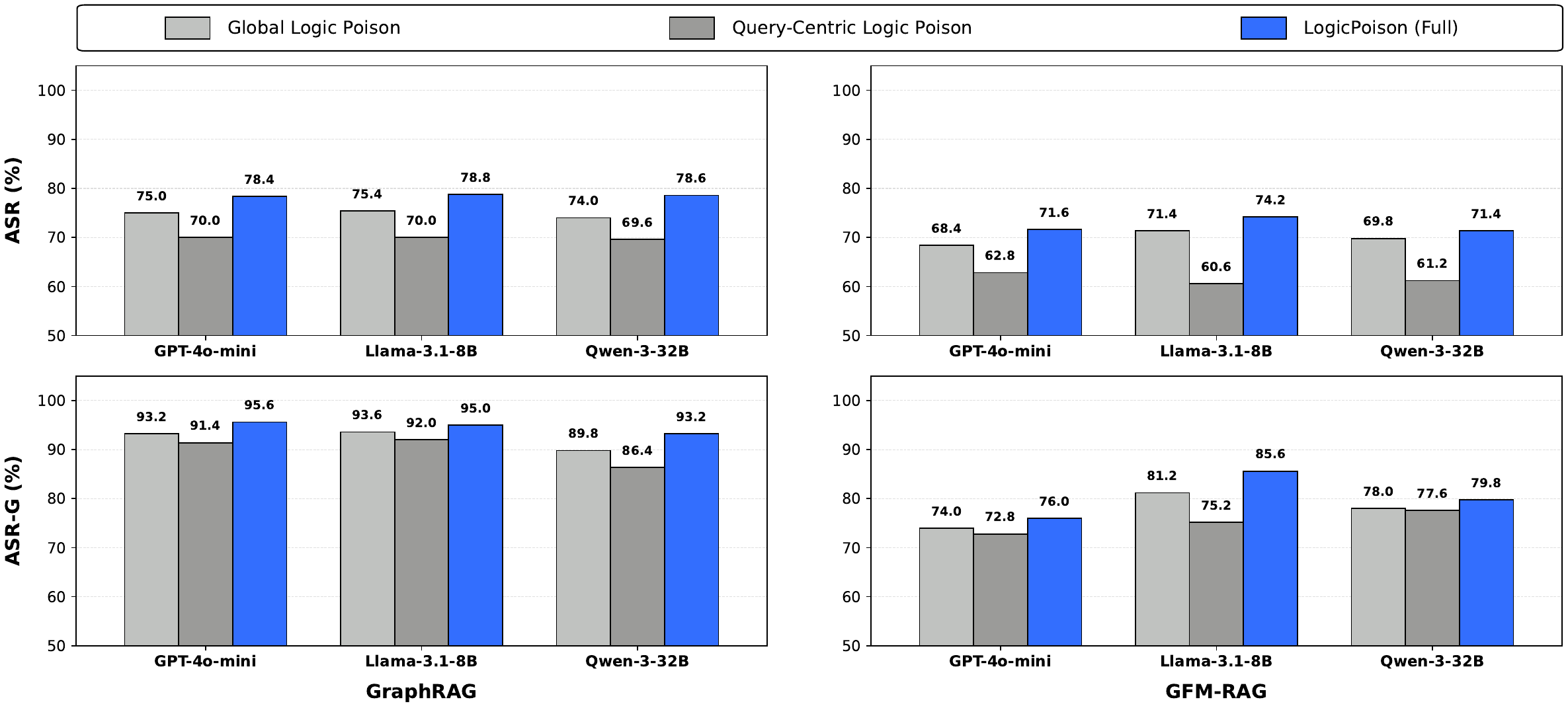}
    \vspace{-3mm}
    \caption{Ablation study of the components on the 2Wiki Dataset. We compare Global-only, Query-Centric-only, and the full \textbf{LogicPoison} strategies with three different LLMs.}
    \label{fig:ablation}
\end{figure*}

To verify the effectiveness of each module, we conducted ablation experiments on three different LLM models and two GraphRAG methods. The experimental results, as shown in Figure ~\ref{fig:ablation}, demonstrate that each module of \textsc{LogicPoison} exhibits significant effectiveness. Specifically, the Query-centric Logic Poison module identifies key entities for each query, effectively cutting off its logical reasoning chain and contributing to a significant attack effect; the Global Logic Poison module, by weakening the Logic Hubs in the Knowledge Graph, severely undermines the overall logicality of the graph which does not rely on LLM or query content, making it the preferred solution in scenarios with low costs and low data availability. The combination of the two forms \textsc{LogicPoison}, which effectively integrates the aforementioned advantages: it not only disrupts the topological connections of the KG at a global level but also specifically interrupts the exclusive logical reasoning chain of each query, achieving the optimal attack effect.

\subsection{Potential Defenses}
Previous experiments have fully demonstrated the effectiveness of \textsc {LogicPoison} in attacking GraphRAG. In the following section, this paper will further explore potential defense mechanisms.

\begin{table}[h]
\centering
\small
\setlength{\tabcolsep}{2.1pt} 
\begin{tabular}{llccc}
\toprule
\multirow{2}{*}{\textbf{Dataset}} & \multirow{2}{*}{\textbf{System}} & \textbf{Standard} & \textbf{w/ Paraphrasing} & \multirow{2}{*}{\textbf{$\Delta$}} \\
 &  & (No Def.) & (Defense) &  \\
\midrule
\multirow{2}{*}{HotpotQA} & GraphRAG & 92.2 & 92.2 & 0 \\
 & GFM-RAG & 78.0 & 77.8 & -0.2 \\
\midrule
\multirow{2}{*}{2Wiki} & GraphRAG & 95.6 & 94.6 & -1.0 \\
 & GFM-RAG & 76.0 & 75.6 & -0.4 \\
\midrule
\multirow{2}{*}{MuSiQue} & GraphRAG & 97.0 &96.2 & -0.8 \\
 & GFM-RAG & 93.4 & 93.4 & 0 \\
\bottomrule
\end{tabular}
\caption{\textbf{Defense Analysis.} Robustness of \textsc{LogicPoison} against Query Paraphrasing defense across different datasets and GraphRAG systems. The LLM backbone is fixed to GPT-4o-mini.}
\label{tab:defense_paraphrasing}
\end{table}

\subsubsection{Query Paraphrasing}

Paraphrasing~\cite{Paraphrasing} is widely used to defend against prompt injection attacks~\cite{injection,Injection2,Injection3,Injection4} and jailbreak attacks~\cite{Jail1,Jail2} by large language models. Its core mechanism involves rewriting the query by LLM and then performing retrieval and generation tasks based on the rewritten query. We verified the defense effect of this strategy against \textsc {LogicPoison}, generating 5 rewritten versions for each query. The experimental results are shown in Table ~\ref{tab:defense_paraphrasing}. The specific results indicate that Paraphrasing has almost no defense effect against \textsc {LogicPoison}. The decline in attack effectiveness caused by this strategy is always less than 1.0\%. The main reason is that regardless of how the sentence structure of the query is rewritten, its core semantics remain unchanged, which will not affect the extraction results of key entities by \textsc {LogicPoison}, and thus has almost no impact on the overall attack effect.

\subsubsection{LLM Knowledge Referencing}
For the scenario of using GraphRAG, LLM itself cannot solve professional questions well, and a specific knowledge base/KG is needed as the knowledge base. At the same time, the prompt of various graphrag will strictly limit the LLM to refer to the content retrieved from the knowledge base/KG during generation. These constraints greatly limit the defense effect of LLM Knowledge Referencing (relying on LLM internal knowledge)~\cite{Fire} against \textsc{LogicPoison}.

\begin{figure}[h]
    \centering
    \includegraphics[width=1.0\linewidth]{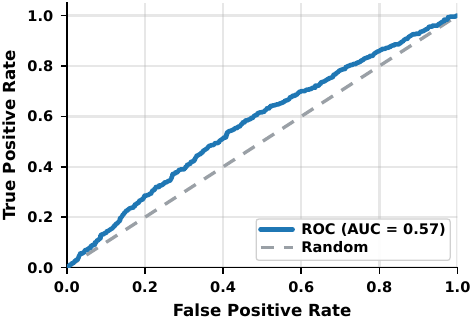}
    \caption{PPL-based detection for \textsc{LogicPoison}.}
    \label{fig:roc}
\end{figure}

\subsubsection{Poisoning Text Identification}
Perplexity (PPL)~\cite{ppl} is widely used to measure text quality and also to defend against attacks on large language models. Previous studies have shown that a high perplexity of text indicates low quality, and poisoned texts often have a higher perplexity than texts written by humans, which makes poisoned texts very easy to detect. To verify this,following ~\cite{poisonedrag}, we calculated the perplexity of clean texts and texts with entities replaced by \textsc{LogicPoison} (randomly sample 500 of each), respectively, using OpenAI's \texttt{tiktoken cl100k\_base model} for detection. As shown in Figure ~\ref{fig:roc}, the results show that the AUC value is 0.57, which is very close to the random guess benchmark of 0.5, indicating that the model is basically in a state of random judgment on poisoned texts. This means that the poisoned texts generated by \textsc{LogicPoison} after entity replacement are almost indistinguishable from clean texts in terms of perplexity, making detection difficult. The core reason lies in the Type-Preserving Entity Swapping module proposed in this paper: all entity replacements maintain the consistency of types, ensuring that the fluency and semantic expression of sentences are not affected, thereby achieving effective evasion of detection mechanisms.

\section{Conclusion}
We found that existing LLM or RAG attack techniques cannot effectively attack the GraphRAG system. Through in-depth analysis of the GraphRAG system, we discovered that it is vulnerable to logic destruction attacks, so we proposed the \textsc{LogicPoison}. This method constructs a unified candidate set by identifying logic hubs and key entities corresponding to queries. Based on cyclic permutation, we implement a comprehensive and covert attack on GraphRAG from both global and query-centric levels, which destroys its logical reasoning ability. Under the experimental settings of three public datasets, three LLM models and three GraphRAG methods, \textsc {LogicPoison} shows excellent performance, and is significantly better than the SOTA in time and token cost. In addition, we explore and experimentally verify the effectiveness of potential defenses. This study provides important guidance for the security research in the GraphRAG field.

\clearpage
\section*{Limitations}
While \textsc{LogicPoison} effectively exposes the topological vulnerabilities of GraphRAG and demonstrates high efficacy across multiple benchmarks, we acknowledge certain limitations that point to directions for future research.

\paragraph{Generalization to Dynamic Graph Construction.} Our experiments focus on graph indices built from static corpora. Real-world GraphRAG systems may employ dynamic updating mechanisms where nodes are added or pruned in real-time. While the theoretical basis of logical poisoning remains valid, the persistence and stability of the attack under high-frequency graph updates warrant further empirical investigation.

\paragraph{Linguistic Scope.} The current evaluation is confined to English-language datasets. Although logical reasoning is language-agnostic, implementing imperceptible entity swapping in languages with complex morphology or rich inflection requires tailored rewriting heuristics. We plan to extend our validation to multilingual GraphRAG settings in future work.

\section*{Ethics Statement}

This work complies with the ACL Ethics Policy. We utilize three publicly available datasets: HotpotQA, 2WikiMultiHopQA, and MuSiQue, which are widely established benchmarks in the research community. To the best of our knowledge, these datasets are free of personally identifiable information (PII) or offensive content. Our study involves neither human subject experimentation nor the collection of private user data.

We acknowledge that the \textsc{LogicPoison} framework proposed in this paper demonstrates a method to compromise GraphRAG systems, which carries a potential risk of misuse. However, our primary motivation is defensive: by exposing the fragility of topological reasoning in GraphRAG, we aim to alert the community to these stealthy vulnerabilities. We believe that identifying such structural weaknesses is a prerequisite for developing more robust defense mechanisms and secure RAG systems. All experiments were conducted in a controlled environment, and no poisoned data was released into real-world applications.

\bibliography{custom}
\clearpage
\appendix

\section{Appendix}
\label{sec:appendix}
\subsection{Analysis of the graph-based architecture}
\label{appendix:Analysis}
Unlike traditional RAG, which relies on similarity in vector space to retrieve static text chunks, GraphRAG fundamentally operates as a topological reasoning engine. Its core advantage does not lie in merely accumulating knowledge triples, but in being able to perform transitive reasoning between these triples.

Specifically, GraphRAG systems synthesize answers using graph traversal algorithms (e.g., multi-hop pathfinding) and structural aggregation mechanisms (e.g., community detection via the Leiden algorithm). In this architecture, the value of a piece of information depends on its connectivity. For example, answering a complex query often requires traversing specific inferential bridges. If a semantic bridge is topologically disconnected or routed incorrectly, the system cannot connect premises to conclusions even if both entities are intact in the database.

Thus, the robustness of GraphRAG depends on structural integrity rather than data volume. This architectural dependency exposes a critical vulnerability: logic is fragile. While vector retrieval degrades moderately under noise (retrieved text chunks are slightly less relevant), GraphRAG's inference chains are prone to catastrophic failures when critical hub nodes or bridge edges are damaged. Replacing a key entity in the inference chain not only introduces factual errors but also disrupts the entire reasoning process, preventing the retrieval mechanism from identifying the correct context. This insight shifts the attack target from injecting large amounts of false information (flooding attacks) to subtly reconstructing logic (reconnection attacks), which also prompts us to propose the \textsc{LogicPoison} framework.
\subsection{Theoretical Analysis of logic hub}
\subsubsection{Hub Nodes and Connectivity Disruption}
Real-world graphs, including social networks, information networks, and knowledge graphs constructed from large-scale corpora, often exhibit scale-free properties. In such networks, node degrees follow a heavy-tailed distribution rather than a homogeneous distribution. Formally, the probability that a node has degree $k$ is given by:
\begin{equation}
P(k) \propto k^{-\gamma}, \quad \gamma \in (2,3),
\end{equation}
where a small fraction of nodes, referred to as hub nodes, possess dominantly high degrees, while most of nodes maintain sparse connectivity.

A fundamental result in network science is that scale-free networks are robust to random failures but highly vulnerable to targeted attacks. Random node removal predominantly affects low-degree nodes and therefore preserves the giant connected component. In contrast, selectively removing or perturbing hub nodes can drastically impair global connectivity. Besides, hub nodes play a central role in shortest-path routing and information flow. Their removal significantly increases the average shortest path length and graph diameter, thereby degrading multi-hop information propagation. Prior theoretical and empirical studies have shown that eliminating only a small fraction of the highest-degree nodes can lead to disproportionate degradation of global network efficiency.

This behavior is commonly analyzed through percolation theory, where the existence of a giant connected component depends on the ratio:
\begin{equation}
\kappa = \frac{\langle k^2 \rangle}{\langle k \rangle}.
\end{equation}
For scale-free networks with $\gamma \leq 3$, the second moment $\langle k^2 \rangle$ diverges, resulting in a vanishing percolation threshold under random failures. However, targeted attacks on hub nodes sharply reduce $\langle k^2 \rangle$, driving the network below the critical threshold and causing abrupt fragmentation.

\subsubsection{Node Centrality Metrics
}

To quantitatively characterize the importance of entities in a graph, we adopt classical centrality measures from network science that capture complementary aspects of node influence. Degree centrality measures the number of direct connections of a node, formally defined as $C_D(v) = \deg(v)$, reflecting its local prominence and immediate aggregation capacity; in GraphRAG-style entity graphs, nodes with high degree centrality typically correspond to frequently mentioned or semantically generic entities that serve as global connectors. In contrast, betweenness centrality quantifies the extent to which a node lies on shortest paths between other nodes, defined as $C_B(v) = \sum_{s \neq v \neq t} \frac{\sigma_{st}(v)}{\sigma_{st}}$, where $\sigma_{st}$ denotes the number of shortest paths between nodes $s$ and $t$ and $\sigma_{st}(v)$ those passing through $v$; such nodes act as structural bridges that mediate information flow and multi-hop reasoning. Finally, closeness centrality, defined as $C_C(v) = \left(\sum_{u \neq v} d(v,u)\right)^{-1}$, captures how close a node is to all others in terms of shortest-path distance, reflecting its ability to rapidly access or disseminate information across the graph. Together, these metrics provide a quantitative basis for identifying nodes that are critical to global connectivity, efficient information propagation, and reasoning path composition.

\subsubsection{Spectral View of Frequent Entities}
\label{sec:spectral}

From the perspective of spectral graph theory, the global connectivity and information propagation properties of a graph are governed by the spectrum of its adjacency or Laplacian matrix. Let $G = (V,E)$ denote the entity graph induced from a corpus, and let $A \in \mathbb{R}^{|V| \times |V|}$ be its adjacency matrix. The largest eigenvalue $\lambda_{\max}(A)$ and the associated leading eigenvector play a dominant role in characterizing diffusion dynamics, aggregation behavior, and global structural coherence.

In heterogeneous graphs, particularly those with scale-free characteristics, the leading eigenvalue is disproportionately influenced by high-degree nodes. Classical results in spectral graph theory indicate that the spectral radius scales with the square root of the maximum degree:
\begin{equation}
\lambda_{\max}(A) \approx \max_{v \in V} \sqrt{\deg(v)},
\end{equation}
suggesting that nodes with large degrees exert a dominant influence on the spectral radius. Moreover, the mass of the leading eigenvector tends to localize around hub nodes, a phenomenon known as eigenvector localization, implying that these nodes dominate global signal propagation.

In corpus-induced entity graphs, explicit degree information is often unavailable prior to graph construction. However, under standard co-occurrence-based graph formation, the expected degree of an entity is monotonically related to its corpus-level frequency. Let $f(e)$ denote the document frequency of entity $e$. 

\begin{equation}
\mathbb{E}[\deg(e)] \propto f(e),
\end{equation}

Consequently, high-frequency entities are expected to dominate the eigenspace of the induced graph, which motivates the use of corpus-level frequency as a practical substitute for spectral dominance. Attacking or perturbing such entities induces a low-rank perturbation to the adjacency matrix. Let $\tilde{A} = A + \Delta A$ denote the perturbed adjacency matrix after poisoning a small set of high-frequency entities. The First-order matrix perturbation theory yields the following.
\begin{equation}
\Delta \lambda_{\max} \approx \mathbf{u}^\top \Delta A \mathbf{u},
\end{equation}
where $\mathbf{u}$ is the leading eigenvector of $A$. Since $\mathbf{u}$ concentrates its mass on hub nodes, perturbations targeting these nodes produce disproportionately large shifts in $\lambda_{\max}$. Such spectral shifts degrade global diffusion efficiency and destroy multi-hop information aggregation in an efficient manner.

Taken together, these results imply that poisoning high-frequency entities constitutes a spectrally efficient attack strategy: by targeting a small number of entities that dominate the leading eigenspace, the attack induces global structural degradation with minimal perturbation budget. This provides a theoretical explanation for why corpus-level frequency-based attacks can effectively disrupt reasoning in GraphRAG systems without requiring explicit access to the full graph topology.

\begin{algorithm}[htbp]
    \caption{\textsc{LogicPoison} Attack Workflow}
    \label{alg:logicpoison}
    \textbf{Input}: Clean Corpus $\mathcal{C}$, Query Set $\mathcal{Q}$, Entity Types $\mathcal{T}$, Poison Budget $n\%$.\\
    \textbf{Output}: Poisoned Corpus $\tilde{\mathcal{C}}$.\\
    \begin{algorithmic}[1]
        \STATE Initialize target pools: $\mathcal{G} \leftarrow \emptyset$, $\mathcal{Q}_{pool} \leftarrow \emptyset$, $\mathcal{R} \leftarrow \emptyset$;
        
        \STATE \COMMENT{\textbf{Phase 1: Global Logic Poisoning}}
        \FOR{document $d_i \in \mathcal{C}$}
            \STATE Extract typed entities $E(d_i)$ using NER;
            \STATE Update frequency counts $f(e, \tau)$;
        \ENDFOR
        \FOR{type $\tau \in \mathcal{T}$}
            \STATE Select top-$n\%$ frequent entities as $\mathcal{G}_\tau$;
        \ENDFOR
        \STATE $\mathcal{G} \leftarrow \bigcup_{\tau} \mathcal{G}_\tau$;

        \STATE \COMMENT{\textbf{Phase 2: Query-Centric Logic Poisoning}}
        \FOR{query $q_j \in \mathcal{Q}$}
            \STATE Extract reasoning chain entities $E_{q_j}$ via LLM CoT;
            \STATE Filter $E_{q_j}$ against corpus inventory to get valid set $E_{q_j}^{\mathrm{corpus}}$;
            \STATE $\mathcal{Q}_{pool} \leftarrow \mathcal{Q}_{pool} \cup E_{q_j}^{\mathrm{corpus}}$;
        \ENDFOR

        \STATE \COMMENT{\textbf{Phase 3: Unified Swapping Mechanism}}
        \STATE $\mathcal{R} \leftarrow \mathcal{G} \cup \mathcal{Q}_{pool}$; \COMMENT{Combine global and query-specific targets}
        \FOR{type $\tau \in \mathcal{T}$}
            \STATE Extract subset $\mathcal{R}_\tau$ from $\mathcal{R}$ where type is $\tau$;
            \STATE Sort $\mathcal{R}_\tau$ to form sequence $\mathbf{S}_\tau$;
            \STATE Construct cyclic permutation function $\pi_\tau$: $\mathbf{S}_\tau[i] \to \mathbf{S}_\tau[i-1]$;
        \ENDFOR

        \STATE \COMMENT{\textbf{Phase 4: Corpus Rewriting}}
        \STATE $\tilde{\mathcal{C}} \leftarrow \emptyset$;
        \FOR{document $d_i \in \mathcal{C}$}
            \STATE $\tilde{d}_i \leftarrow d_i$;
            \FOR{entity $e \in \mathcal{R}$}
                \STATE Replace all mentions of $e$ in $\tilde{d}_i$ with $\pi_{\text{type}(e)}(e)$;
            \ENDFOR
            \STATE $\tilde{\mathcal{C}}$.append$(\tilde{d}_i)$;
        \ENDFOR
        
        \STATE \textbf{return} Poisoned Corpus $\tilde{\mathcal{C}}$.
    \end{algorithmic}
\end{algorithm}

\subsection{Algorithm Workflow}

Algorithm \ref{alg:logicpoison} details the execution pipeline of the \textsc{LogicPoison} framework. The process operates in three phases: (1) Target Selection, where global logic hubs and query-centric bridges are identified; (2) Permutation Construction, where a type-preserving cyclic mapping is defined for the unified target set; and (3) Corpus Rewriting, which generates poisoned corpus $\tilde{\mathcal{C}}$ to induce graph corruption.

\subsection{Details of GraphRAG methods}
Here we introduce the graphrag methods chosen in this paper in detail:

\paragraph{GraphRAG.}Microsoft GraphRAG proposes a hierarchical community summarization strategy based on knowledge graph. This method constructs a knowledge graph by extracting entities and relations through LLM, and then generates a structured summary through hierarchical community detection. Finally, map-reduce mode is used to integrate partial responses of each community to generate a global answer. The innovation of this method is that it combines the modularity of graph with summary generation, and supports global queries on millions of token corpora. It is significantly better than the traditional vector RAG in the comprehensiveness and diversity of answers, and provides an efficient solution for the global sense-making task of large-scale private corpora.

\paragraph{GFM-RAG.}In order to solve the problems of noisy graph structure, incompleteness and lack of model generality in GraphRAG, GFM-RAG proposes the first Graph-based model (GFM) suitable for retrieval enhancement generation. The model takes query dependency graph neural network as the core, and through two-stage training (unsupervised knowledge graph completion pre-training and supervised document retrieval fine-tuning), it learns general reasoning patterns on large-scale data containing 60 knowledge graphs and 700k documents, and can directly adapt to unseen data sets without fine-tuning. Its innovative single-step multi-hop inference mechanism significantly improves the performance of complex reasoning tasks while ensuring efficiency, and conforms to the neural scaling law, which provides a basis for subsequent model expansion.

\paragraph{HippoRAG 2.}Aiming at the problem of performance degradation of existing structure enhanced RAG methods in fact memory tasks, HippoRAG 2 is based on a neurobiologically inspired long-term memory framework and inherits the core of HippoRAG's personalized PageRank algorithm. Through the optimization of deep paragraph integration, deep association between query and knowledge graph triples, and LLM online identification of irrelevant triples, the problem of context loss and semantic matching caused by the entity-centric method is solved.

We have carefully selected these three most representative graphrag methods as part of the experiment. Due to the huge resource consumption of graphrag methods, we were unable to include more other excellent graphrag methods. In future, we will consider expanding the scope of the experiments.


\subsection{Details of Baselines}
\label{appendix_baselines}
Here we introduce the baselines chosen in this paper in detail:

\paragraph{PoisonedRAG.}PoisonedRAG is the first knowledge pollution attack framework against RAG. This work reveals for the first time that the knowledge base of RAG systems can be used as a practical attack surface to induce LLMS to generate incorrect answers preset by attackers for specific target questions by injecting a small amount of malicious text. Its core innovation lies in deriving and satisfying two necessary conditions for the attack: retrieval condition (malicious text can be retrieved by the target question) and generation condition (malicious text can mislead the LLM to output the target answer).

\paragraph{Naive Attack.}Following the baseline~\cite{poisonedrag}, we define Naive Attack as: Given a question, treat the question itself as malicious text because it has a high probability of being retrieved.

\paragraph{Prompt Injection.} Prompt Injection~\cite{injection} is a kind of security attack against Large Language Model integration applications. The core of prompt injection is that the attacker can inject malicious instructions or data by manipulating the external input data of the application, and induce the LLM to ignore the original task instructions and execute the injection task preset by the attacker, so as to produce the attacker's expected results.

\paragraph{GCG Attack.} By combining greedy search and gradient-directed discrete optimization strategies, GCG Attack~\cite{GCG} automatically generates adversarial suffixed, which can efficiently induce aligned LLMS to output harmful content. The core innovations of this method are as follows: first, the model is guided into harmful generation mode by optimizing the target text prefix without manual intervention; Second, it supports multi-hint and multi-model joint optimization, and the generated adversarial suffix has strong versatility and transferability, which can successfully attack GPT-4, Bard and other black-box commercial models and open source LLMs.

\paragraph{Disinformation.} PoisonedRAG generates the basis content of the attack text through the prompt designed for the target problem, which can be regarded as disinformation~\cite{Disinformation}. Therefore, we compare with this baseline, which can be seen as a variant of PoisonedRAG.

\begin{figure}[h]
    \centering
    \includegraphics[width=1.0\linewidth]{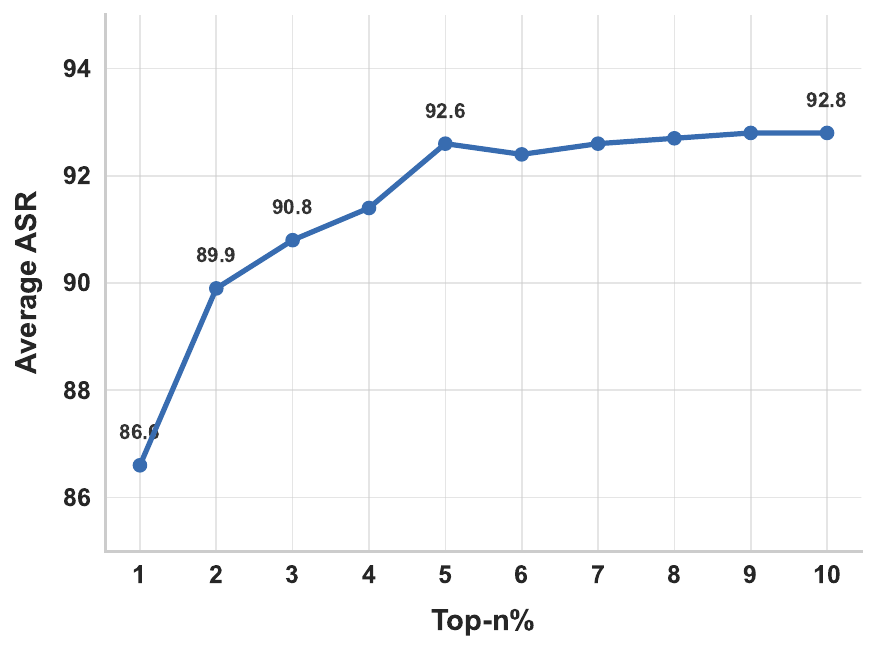}
    \caption{Hyperparameter analysis results of \textsc{LogicPoison}.}
    \label{fig:hyper}
\end{figure}

\subsection{Hyperparameter analysis}

\textsc{LogicPoison} contains only one hyperparameter, which is the top-n\% in the Global Logic Poison module. To select the smallest value that balances effectiveness and efficiency, we conducted a hyperparameter sensitivity analysis based on Naive RAG(since the cost of repeating experiments multiple times on the GraphRAG method is extremely high) under the setting of retrieval top-k=5. All Attack Success Rate results are the average of 5 independent tests on 3 datasets. The experimental results show that when n\% exceeds 5\%, the improvement in ASR tends to level off. Therefore, we finally chose top-n\%=5\% as the hyperparameter, which achieves the optimal balance between attack effect and resource cost.

\subsection{Prompts}

In this section, we present the prompt templates that drive the components of our method. Figure \ref{prompts_entity_extraction} illustrates the entity extraction prompt, which enforces strict constraints to isolate minimal reasoning hops and classify entities (including implicit bridges). Additionally, Figure \ref{prompts_exact_match} shows the evaluation prompt used to verify the ASR-GPT of the final answers, ensuring robustness against formatting variations.

\begin{figure*}[h]
\begin{tcolorbox}[colback=gray!10, colframe=LightBlue , title= ASR-GPT Evaluation Prompt]
\small
You are an expert judge evaluating if a model's prediction exactly matches the correct answer.

\vspace{10pt}

\textbf{Question:}

\textless Question\textgreater

\vspace{10pt}

\textbf{Prediction:}

\textless Prediction\textgreater

\vspace{10pt}

\textbf{Correct answer:}

\textless Correct Answer\textgreater

\vspace{10pt}

\textbf{Task:} Determine if the prediction exactly matches the correct answer. Consider:
\begin{itemize}[noitemsep, topsep=0pt, leftmargin=1.5em]
    \item The prediction must contain the same core information as the correct answer.
    \item Minor formatting differences (punctuation, capitalization, spacing) should be ignored.
    \item The prediction should convey the same meaning as the correct answer.
\end{itemize}

\vspace{10pt}

Respond with only "YES" or "NO" (no other text).
\end{tcolorbox}
\caption{Prompt template used for evaluating exact matches between predictions and gold answers.}
\label{prompts_exact_match}
\end{figure*}

\begin{figure*}[h]
\begin{tcolorbox}[colback=gray!10, colframe=LightBlue, title= Entity Extraction Prompt]
\small
Extract the reasoning-critical entities from the multi-hop question Q.

\vspace{5pt}
\textbf{Your constraints:}
\begin{itemize}[noitemsep, topsep=0pt, leftmargin=1.5em]
    \item Do NOT answer Q.
    \item Do NOT add external knowledge.
    \item Use ONLY the wording of Q.
    \item Identify the minimal reasoning hops (1, 2, 3\dots).
    \item For each hop, extract only entities necessary for the reasoning chain.
\end{itemize}

\vspace{5pt}
\textbf{Entity types:}
Use spaCy's NER label set when applicable:
``PERSON'', ``NORP'', ``FAC'', ``ORG'', ``GPE'', ``LOC'', ``PRODUCT'', ``EVENT'', ``WORK\_OF\_ART'', ``LAW'', ``LANGUAGE'', ``DATE'', ``TIME'', ``PERCENT'', ``MONEY'', ``QUANTITY'', ``ORDINAL'', ``CARDINAL''.

Additionally allow two reasoning-specific types:
\begin{itemize}[noitemsep, topsep=0pt, leftmargin=1.5em]
    \item ``ALIAS'': paraphrased or descriptive label referring to an entity (e.g., ``the nation called the nobilities commonwealth'')
    \item ``BRIDGE'': implicit descriptive entity needed to connect reasoning hops (e.g., ``the country where the operatives originated'')
\end{itemize}

\vspace{5pt}
\textbf{Guidelines:}
\begin{itemize}[noitemsep, topsep=0pt, leftmargin=1.5em]
    \item Include implicit bridge entities (BRIDGE) and alias-style descriptions (ALIAS) if they are part of the reasoning chain.
    \item Do NOT normalize entities to real-world names.
    \item Focus only on entities whose presence or value matters for understanding or resolving the question.
\end{itemize}

\vspace{5pt}
\textbf{Output format (strict):}
A Python-style list of dicts, e.g.:
\begin{verbatim}
[
  {"hop": 1, "entity": "...", "type": "GPE"},
  {"hop": 2, "entity": "...", "type": "BRIDGE"},
  {"hop": 3, "entity": "...", "type": "DATE"}
]
\end{verbatim}
Return ONLY the list.

\vspace{5pt}
\textbf{Q:} \textless Question Text\textgreater

\end{tcolorbox}
\caption{Prompt template for reasoning-critical entity extraction.}
\label{prompts_entity_extraction}
\end{figure*}

\end{document}